\documentclass{article} 
\usepackage{iclr2024_conference,times}

\usepackage{amsmath,amsfonts,bm}

\def\eqref#1{equation~\ref{#1}}

\def\1{\bm{1}}

\def\vzero{{\bm{0}}}

\def\vx{{\bm{x}}}

\def\mI{{\bm{I}}}

\DeclareMathAlphabet{\mathsfit}{\encodingdefault}{\sfdefault}{m}{sl}
\SetMathAlphabet{\mathsfit}{bold}{\encodingdefault}{\sfdefault}{bx}{n}

\def\gN{{\mathcal{N}}}

\usepackage{hyperref}
\usepackage{url}
\usepackage{graphicx}
\usepackage{enumitem}
\usepackage{svg}
\usepackage{wrapfig}
\usepackage{booktabs}
\usepackage{bbding}
\usepackage{utfsym}
\usepackage{array}
\usepackage{xcolor}
\usepackage{colortbl}
\usepackage{pifont}

\newcolumntype{C}[1]{>{\centering\arraybackslash}p{#1}}

\newcommand{\Mat}{\boldsymbol}

\DeclareMathOperator{\mean}{\mathbb{E}}

\title{Efficient-3Dim: Learning a Generalizable Single-image Novel-view Synthesizer in One Day}

\author{Yifan Jiang\textsuperscript{1,2}\thanks{This work was performed while Yifan Jiang interned at Apple.}, Hao Tang\textsuperscript{1}, Jen-Hao Rick Chang\textsuperscript{1}, Liangchen Song\textsuperscript{1},\\ \textbf{Zhangyang Wang\textsuperscript{2}, Liangliang Cao\textsuperscript{1}}\\
\textsuperscript{1}Apple, \textsuperscript{2}University of Texas at Austin\\
\texttt{\{yifanjiang97, atlaswang\}@utexas.edu} \\
\texttt{\{hao\_tang,jenhao\_chang,liangchen\_song,llcao\}@apple.com} \\
}

\iclrfinalcopy 
\begin{document}

\maketitle

\begin{abstract}
The task of novel view synthesis aims to generate unseen perspectives of an object or scene from a limited set of input images. Nevertheless, synthesizing novel views from a single image still remains a significant challenge in the realm of computer vision. Previous approaches tackle this problem by adopting mesh prediction, multi-plain image construction, or more advanced techniques such as neural radiance fields. Recently, a pre-trained diffusion model that is specifically designed for 2D image synthesis has demonstrated its capability in producing photorealistic novel views, if sufficiently optimized on a 3D finetuning task. Although the fidelity and generalizability are greatly improved, training such a powerful diffusion model requires a vast volume of training data and model parameters, resulting in a notoriously long time and high computational costs.
To tackle this issue, we propose \textbf{Efficient-3DiM}, a simple but effective framework to learn a single-image novel-view synthesizer. 
Motivated by our in-depth analysis of the inference process of diffusion models, we propose several pragmatic strategies to reduce the training overhead to a manageable scale, including a crafted timestep sampling strategy, a superior 3D feature extractor, and an enhanced training scheme. When combined, our framework is able to reduce the total training time from 10 days to less than \textbf{1 day}, significantly accelerating the training process under the same computational platform (one instance with 8 Nvidia A100 GPUs). Comprehensive experiments are conducted to demonstrate the efficiency and generalizability of our proposed method. 
\end{abstract}

\section{Introduction} 
Enabling immersive free-viewpoint interaction, especially from a single image, has consistently been a focal point of interest and an intriguing application in the domains of both computer vision and computer graphics. 
To provide realistic interactive imagery, one can initiate by creating an explicit 3D structure (E.g.,  mesh or point cloud) from the existing image and then render the novel viewpoint. Nonetheless, it often lacks view-dependent effects and looks artificial. Since then, neural rendering techniques, such as neural radiance field (NeRF~\cite{mildenhall2020nerf}) and scene representation networks (SRN~\cite{sitzmann2019scene}), have prevailed over all other approaches and become the crucial step in reaching interactivity and realism. Instead of constructing an explicit representation, these approaches learn an implicit representation by modeling a 3D scene as a continuous function.

Although NeRF-like models produce compelling details, they are impeded by a stringent requirement---the vanilla NeRF approach requires hundreds of posed images for training and generates photorealistic images on both rendered objects and real-world scenes. Some following works reduce NeRF's training views to as few as three views, by using geometry regularization~\citep{niemeyer2022regnerf}, Structure-from-Motion (SfM) initialization~\citep{deng2022depth}, or adaptive positional encoding~\citep{yang2023freenerf}. 
However, these approaches cannot be easily extended to support a single input view. 
\citet{yu2021pixelnerf} first propose to train a NeRF that can render images from novel viewpoints using a single input, yet it only works on simulated scenes with simple geometry, e.g., ShapeNet~\citep{chang2015shapenet}. \citet{xu2022sinnerf} further extends it to in-the-wild scenes and enables novel view rendering on arbitrary objects. However, it only renders views from a small range of angles.

%
Recently, \citet{watson2022novel} propose the 3DiM pipeline that treats the single-image novel view synthesis as a conditional generative process. By training a diffusion model conditioned on an input image together with the target camera pose relative to the input image, they are able to synthesize novel views of simple synthetic objects in high fidelity.  
\citet{liu2023zero} further extend the method to real-world objects by finetuning a text-to-image model to learn geometry priors using 3D simulated objects. Their proposed zero 1-to-3 method is able to perform zero-shot novel view synthesis in an under-constrained setting, synthesizing view for any in-the-wild image from any angle.

Despite their capability to generate high-fidelity images, training diffusion models for single-image novel view synthesis requires a notoriously long time and prohibitive computational costs.
%
Due to the inherent difficulties of inferring unseen parts in the input image, training requires a large amount of data and a large model capacity.
For example, the 3DiM model~\citep{watson2022novel} contains 471M parameters; the zero 1-to-3 method \citep{liu2023zero} uses a 1.2B-parameter Stable Diffusion model~\citep{rombach2022high} and 800k simulated objects from the Objaverse dataset~\citep{deitke2023objaverse}. 
Training the model requires 10 days on a single instance with eight Nvidia A100 GPUs.
In comparison, training typical image classifiers~\citep{he2016deep,xie2017aggregated,huang2017densely} on the ImageNet dataset~\citep{deng2009imagenet} only takes about \underline{1 day} on a similar platform. 
A straightforward way to shorten the training procedure of diffusion models is to adopt a larger batch size, however, it further increases the required computational resources.
Given the importance of diffusion modes and their applications in 3D vision, an enhanced training approach would greatly aid researchers in hastening the development process, consequently propelling the advancement of this direction.

The major goal of this work is to trim down the training time without spending more costs on the total training resources (e.g., taking large-batch via a distributed system). We present \textbf{Efficient-3DiM}, a refined solution developed to enhance the training efficiency of diffusion models for single-image novel view synthesis. This paper delves into the details and design choices of the framework. Our approach builds upon three core strategies: a revised timestep sampling method, the integration of a self-supervised vision transformer, and an enhanced training paradigm. Together, these elements contribute to a noticeable improvement in training efficiency. Our model, with the proposed enhancements, requires only a single day to train; compared to the original zero 1-to-3, we achieve a 14x reduction in training time in the same computation environment.
The significantly reduced training time enables future research to be rapidly iterated and innovated.

Our major contributions are encapsulated as follows:
\begin{itemize}[leftmargin=*]
    \item We introduce a novel sampling strategy for diffusion timesteps, deviating from the standard uniform sampling, and offering optimized training.
    \item Our framework integrates a self-supervised Vision Transformer, replacing the conventional CLIP encoder, to better concatenate high-level 3D features. 
    \item Combined with an enhanced training paradigm, the proposed Efficient-3DiM reduces the training time from 10 days to less than 1 day, greatly minimizing the training cost to a manageable scale.

\end{itemize}





\section{Related Work}
\paragraph{Diffusion Model.}
Recent breakthroughs in diffusion models~\citep{sohl2015deep,song2019generative,ho2020denoising} have displayed impressive outcomes in generative tasks. These models, as advanced generative tools, produce captivating samples using a sequential denoising method. hey introduce a forward mechanism that infuses noise into data sets and then reverses this mechanism to restore the initial data. Following this, numerous research efforts~\citep{karras2022elucidating,saharia2022photorealistic, dhariwal2021diffusion, nichol2021glide,rombach2022high,  ramesh2022hierarchical,chen2023importance} have been directed toward enhancing the scalability of diffusion models and speeding up the sampling process for improved efficiency. Notably, the LDM model (also known as Stable Diffusion~\citep{rombach2022high} ) stands out, minimizing computational demands by executing the diffusion method on a lower-resolution latent space, making the training of a text-to-image model scalable to the billion-scale web data. 
 Several other investigations have also adapted the diffusion methodology to diverse fields, such as music creation~\citep{huang2023noise2music}, reinforcement learning~\citep{wang2022diffusion}, language generation~\citep{li2022diffusion}, text-to-3D object~\citep{poole2022dreamfusion}, novel view synthesis~\citep{watson2022novel,xu2022neurallift}, video generation~\citep{blattmann2023align}, and more.
In addition to generation, newer research~\citep{SDEdit, zhang2023adding,hertz2022prompt,mokady2022null,brooks2022instructpix2pix,parmar2023zero,goel2023pair,khachatryan2023text2video} also extends the capabilities of diffusion models to tasks related to image and video editing.


\paragraph{Efficient Training for Generative Models.}
Existing research efforts have been undertaken to explore various aspects including data efficiency, model efficiency, training-cost efficiency, and inference efficiency. To address the mode collapse issue under the setting of limited training data, ~\cite{zhao2020differentiable} propose a differentiable augmentation strategy that only takes 10\% training data to train a Generative Adversarial Network (GAN)~\cite{goodfellow2014generative} to achieve similar performance.
Nevertheless, it is unclear whether this method achieves a shorter training time. In addition, since low-resolution images are typically more available than their high-resolution counterparts, ~\citet{chai2022any} propose an any-resolution framework. This framework merges data of varied resolutions to facilitate GAN training, lessening reliance on high-resolution input. However, its impact on training costs and its adaptability for diffusion models have not been thoroughly examined. Another line of research addresses the memory issue during the training stage. ~\citet{lin2019coco} was the pioneer in integrating the patch-wise training approach into GAN frameworks. But due to their discriminator having to amalgamate multiple generated patches, its memory-saving efficiency is not significant. Building on this, \citet{lee2022memory} applied the patch-wise training to the implicit representation framework~\cite{skorokhodov2021adversarial}. While this method does conserve GPU memory, it compromises the quality of the generated samples. Several researchers have sought to mitigate the substantial training and inference expenses tied to diffusion models. \citet{rombach2022high}, for instance, opted to apply diffusion in a latent space, bypassing the pixel space, which effectively cut down both training and inference costs. Other studies, like those of \citep{lu2022dpm,lu2022dpmpp,song2020denoising,bao2022analytic,bao2022estimating}, delved into rapid sampling techniques during inference. However, these do not inherently accelerate the training phase.

\paragraph{Novel View Synthesis from a Single Image.}
Over recent years, the challenge of deriving 3D worlds from a singular image has gained traction. Traditional methods tackled this problem by first employing a monocular depth estimator~\cite{ranftl2020towards}) to predict the 3D geometry and then utilized multi-plane images (MPI~\citep{shih20203d,jampani2021slide}) or point clouds~\cite{mu20223d} to craft artistic renderings. Furthermore, any gaps or inconsistencies seen from new viewpoints were rectified using a previously trained advanced neural network~\cite{yu2019free}. Yet, despite their efficacy, these methodologies have shortcomings. The depth estimator can be unstable, and the rectification techniques might introduce flaws that detract from the realism of the images.
Following approaches~\citep{yu2021pixelnerf,lin2022vision} propose to learn 3D priors from the 3D objects dataset and directly predict a 3D representation from a single input during the inference time. However, these methods face significant quality degradation when processing real-world images, attributable to domain differences. 
~\cite{xu2022sinnerf} proposes to generate novel views by training a neural radiance field that only takes a single RGB-D image. Nevertheless, it needs high-quality depth as the input and only renders new views from a small range of angles. Following works~\citep{xu2023neurallift,tang2023make} get rid of depth input by adopting a guided diffusion loss from a pre-trained text-to-image model. Although these methods are able to render high-fidelity novel views from 360 degrees, they generally require per-scene training, which takes more than several hours. 

More recently, \cite{watson2022novel} propose to solve single-image novel view synthesis by using a conditional generative model. They train a diffusion model as an image-to-image translator, using the current view as the input and predicting a novel view from another angle. To extend this pipeline to in-the-wild images, ~\cite{liu2023zero} choose to borrow the prior knowledge from a 2D text-to-image model~\cite{rombach2022high}. Several other works~\citep{liu2023one,shi2023mvdream,liu2023syncdreamer} also consider multi-view information to enable better 3D consistency. Our proposed method mainly follows this line of works~\citep{watson2022novel,liu2023zero}, but significantly reduces its training cost without performance degradation.

\section{Method}

\subsection{Preliminaries}
\paragraph{Diffusion Models.} Denoising Diffusion Probability Models, or simply called diffusion models, are a class of generative models that learn to convert unstructured noise to real samples.
It produces images by progressively reducing noise from Gaussian noise $p\left(\mathbf{x}_T\right)=\mathcal{N}(\mathbf{0}, \mathbf{I})$, reshaping it to match the target data distribution. The forward diffusion step, represented by $q(\Mat{x}_{t} | \Mat{x}_{t-1})$, introduces Gaussian noise to the image $\Mat{x}_t$. The marginal distribution can be written as: $q\left(\Mat{x}_t \mid \Mat{x_0}\right)=\mathcal{N}\left(\alpha_t \Mat{x_0}, \sigma_t^2 \Mat{I}\right)$, where $\alpha_t$ and $\sigma_t$ are designed to converge to $\mathcal{N}(\Mat{0}, \Mat{I})$ when $t$ is at the end of the forward process~\citep{kingma2021variational,song2020score}. 
In the reverse process $p(\Mat{x}_{t-1} | \Mat{x}_t)$, diffusion models are designed as noise estimators $\Mat{\epsilon}_\theta(\Mat{x}_t, t)$ taking noisy images as input and estimating the noise. Training them revolves around optimizing the weighted evidence lower bound (ELBO)~\citep{ho2020denoising,kingma2021variational}:
\begin{equation}
\mean\left[w(t)\left\|\Mat{\epsilon}_\theta\left(\alpha_t \Mat{x}_0+\sigma_t \Mat{\epsilon} ; t\right)-\Mat{\epsilon}\right\|_2^2\right],
\label{eq:ddpm}
\end{equation}
where  $\Mat{\epsilon}$ is drawn from $\mathcal{N}(\mathbf{0}, \mathbf{I})$, the timestep $t$ follows an unifrom sampling $\mathcal{U}(\mathbf{1}, \mathbf{1000})$, and $w(t)$ serves as a weighting function with $w(t) = 1$ showing impressive results. 
In the inference stage, one can opt for either a stochastic~\citep{ho2020denoising} or a deterministic approach~\citep{song2020denoising}. By selecting $\Mat{x}_T$ from $\mathcal{N}(\mathbf{0}, \mathbf{I})$, one can systematically lower the noise level, culminating in a high-quality image after iterative refinement.

\begin{figure}[!t]
\begin{center}
    \centering
    \includegraphics[width=0.99\linewidth]{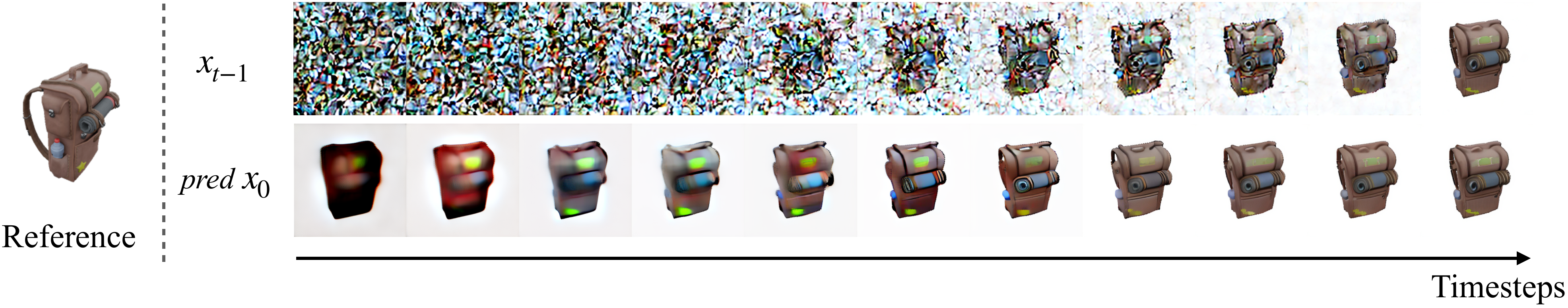} 
\vspace{-1em}
\caption{Inference visualization of a typical diffusion-based novel-view synthesizer~\citep{liu2023zero}. We run a total of 200 steps following Eq.~\ref{eq:ddim} and show the intermediate inference progress of a ``backpack'' under different timesteps.}
\label{fig:visualization}
\end{center}
\end{figure}
\paragraph{Diffusion Model as Novel-view Synthesizer.} 
The standard diffusion model serves as a noise-to-image generator, yet a modified one can be extended to an image-to-image translator, using a conditional image as the reference. By adopting a pair of posed images $\{x_0, \widehat{x_0}\} \in \mathbb{R}^{H \times W \times 3}$ from the same scene for training, an image-to-image diffusion model can take the image, $\widehat{x_0}$, as the input conditioning to predict the image from a different view, $x_0$, approximating a single-image novel-view synthesizer~\citep{watson2022novel,liu2023zero}. Specifically, the training objective of diffusion models becomes:
\begin{equation}
\mean\left[w(t)\left\|\Mat{\epsilon}_\theta\left(\alpha_t \Mat{x}_0+\sigma_t \Mat{\epsilon} ; t ; C\left(\widehat{x_0}, R, T\right)\right)-\Mat{\epsilon}\right\|_2^2\right],
\label{eq:ddpm}
\end{equation}
where $C$ is a feature extractor, $\{R, T\}$ are the relative rotation and translation between $\widehat{x_0}$ and $x_0$. 
%
Since the task of single-image novel view synthesis is severely under-constrained, the training requires a huge dataset consisting of diverse real 3D objects. The current largest open-sourced 3D dataset Objaverse~\citep{deitke2023objaverse} only contains 800k synthetic objects, largely behind the 5 billion in-the-wild yet annotated 2D image dataset~\citep{schuhmann2022laion}, not to mention the domain gap between synthetic and real samples.
%
%
\citet{liu2023zero} proposes Zero 1-to-3 by initializing from a pre-trained 2D text-to-image diffusion model, successfully generalizing to real scenarios with in-the-wild images. Although Zero 1-to-3 diminishes the need for 3D training data and reduces the training cost as well, it still requires a laborious effort to converge. For instance, the official training scheme requires 10 days to run on 8 Nvidia-A100 GPUs.
Our proposed method is built on top of the framework but further reduces its training time to a manageable scale.

\begin{figure*}[!t]
\vspace{-1em}
\begin{center}
    \centering
    \includegraphics[width=0.99\linewidth]{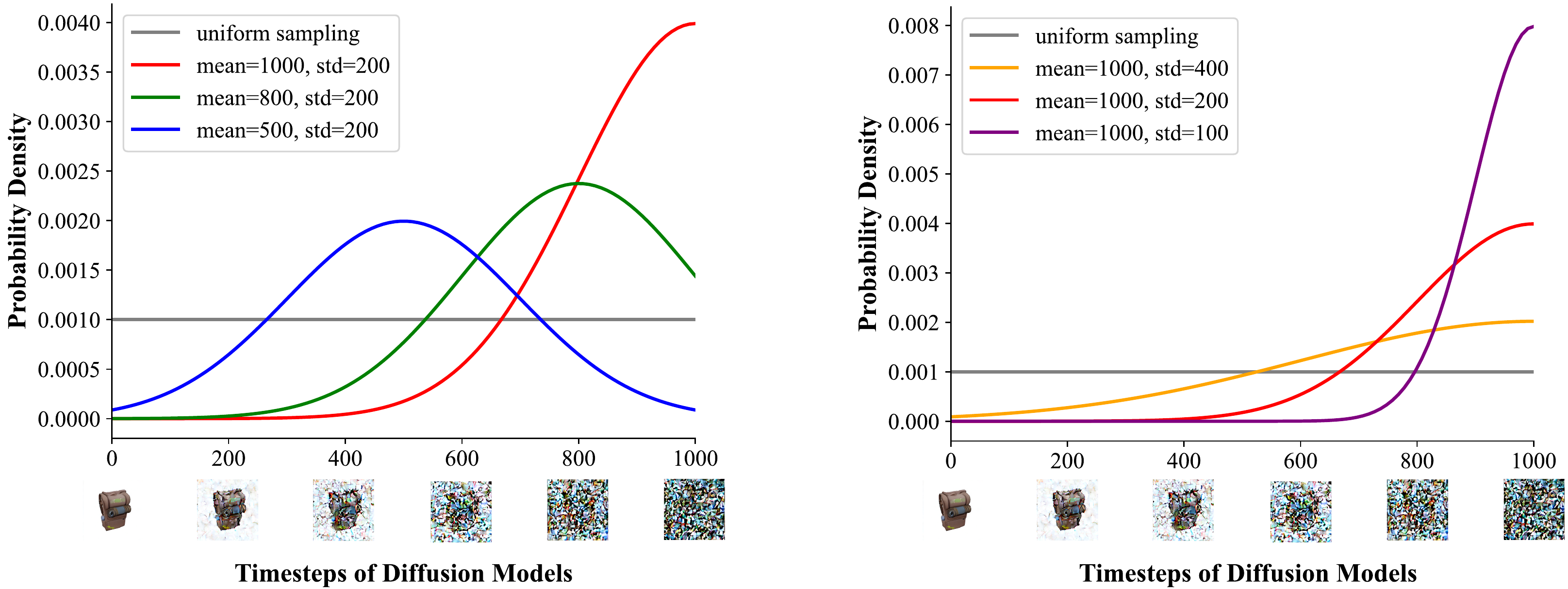} 
\caption{Gaussian sampling under different $mean$ and $std$ values. The left figure shows how the $mean$ factor could control the sampling bias towards different stages of timestep $t$, and the right figure shows how much the bias could be applied via controlling the value of $std$. The distribution is adjusted to ensure the integral of the probability density function ranges from [0, 1000] is 1. }
\label{fig:gaussian}
\end{center}
\vspace{-1.em}
\end{figure*}

\subsection{Modified Sampling Strategies}
\label{sec:sampling_strategy}
We start by analyzing the inference progress of the current state-of-the-art novel-view synthesizer~\citep{liu2023zero}. 
First, we adopt denoising diffusion implicit models \citep{song2020denoising}:
\begin{align}
\label{eq:ddim}
    & \vx_{t-1} = \sqrt{\alpha_{t-1}} \underbrace{\left(\frac{\vx_t - \sqrt{1 - \alpha_t} \cdot \Mat{F}}{\sqrt{\alpha_t}}\right)}_{\text{`` predicted } \vx_0 \text{''}} + \sqrt{1 - \alpha_{t-1} - \sigma_t^2} \cdot \Mat{F}) + \sigma_t \epsilon_t \\
    & \Mat{F}  = \Mat{\epsilon}_\theta(\vx_t ; t; C\left(\widehat{\vx_0}, R, T\right)),
\end{align}
where $\epsilon_t \sim \gN(\vzero, \mI)$ is standard Gaussian noise, and we let $\alpha_0$ to be $1$. Different choices of $\sigma$ values result in different generative processes, under the same model $\epsilon_\theta$. When $\sigma_t = 0$, the generative process degrades to a deterministic procedure and the resultant framework becomes an implicit probabilistic model~\citep{mohamed2016learning}. Here we do not pay attention to the stochasticity of the inference process. Instead, we analyze the output of $\epsilon_\theta$ --- the ``predicted $\vx_0$'', to better understand the learned 3D knowledge. 

From Figure~\ref{fig:visualization}, we observe that the structure and the geometry of the predicted ``backpack'' are constructed in the early stage of the reverse process, while the color and texture are refined in the late stage. Note that the adopted novel-view synthesizer is initiated from a pre-trained text-to-image diffusion model, the knowledge of refining texture details should be also inherited. 
This observation prompts us to reconsider the prevailing uniform sampling methodology used for the diffusion timestep in the training stage. Instead, we advocate for the implementation of a meticulously designed Gaussian sampling strategy.

It is worth mentioning that the potential usage of different sampling strategies has actually been explored before~\citep{chen2023importance,karras2022elucidating}. However, our work distinguishes itself, due to the fact that the major phase of our framework's training is essentially characterized as a finetuning paradigm. This stands in contrast to the predominant works which generally focus on building diffusion models from scratch.
When juxtaposed with the uniform sampling approach, Gaussian sampling offers a distinct advantage: it permits the introduction of a sampling bias, thus conserving efforts that might otherwise be expended on superfluous segments of the training.
We show several typical Gaussian sampling strategies with different mean and std in Figure~\ref{fig:gaussian}. By adopting such a different sampling strategy, we are able to adjust the sample probability under different timesteps. 
Consequently, the corresponding timesteps that are still in development should receive augmented opportunities for updates, while the other parts get a lower probability, but not zero probability. From the distribution visualization, some of these Gaussian distributions satisfy our goal, such as \{$mean = 1000$ and $std = 200$\} as depicted in Figure~\ref{fig:gaussian}. Adopting a slightly different hyperparameter may also reach a similar goal, where we simply choose an effective one by empirical observation.
We show a detailed experiment to demonstrate its effectiveness in Section~\ref{sec:exp_gaussian}

\begin{figure*}[!t]
\vspace{-1em}
\begin{center}
    \centering
    \includegraphics[width=0.8\linewidth]{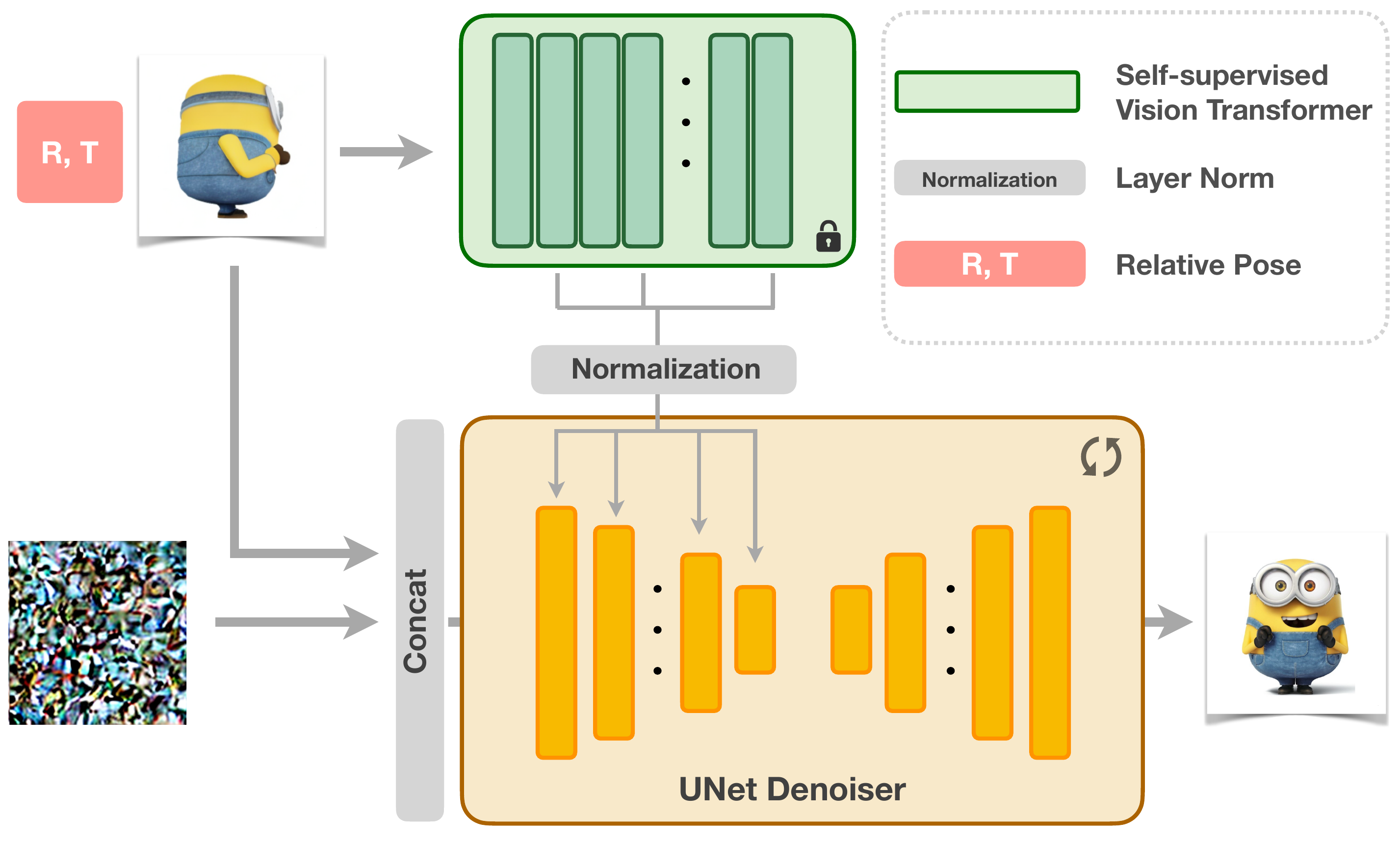} 
\caption{\textbf{Main inference pipeline of Efficient-3DiM framework. }
We finetune the text-to-image diffusion model (Stable Diffusion~\citep{rombach2022high}) on the Objaverse dataset~\citep{deitke2023objaverse} but replace the original CLIP encoder with a self-supervised vision transformer (DINO-v2~\citep{oquab2023dinov2}). Rather than only adopting the ``[CLS]'' token from the the output of reference feature extractor, we amalgamate multi-scale representations into different stages of the UNet denoiser.}
\label{fig:pipeline}
\end{center}
\vspace{-2.em}
\end{figure*}

\subsection{3D Representation Extraction and Amalgamation}
Within the context of the aforementioned discussion, the task of predicting a novel view from a single input is highly ill-posed, as a singular 2D image could be projected from totally different 3D representations. Consequently, the optimization of diffusion models is expected to ensure that the 2D reference could be back-projected to the latent 3D space that aligned with human perception. Predominant works either directly concatenate the 2D condition with original noisy input and train the denoiser to capture the 3D geometry~\citep{watson2022novel}, or integrate an auxiliary CLIP image encoder~\citep{radford2021learning} to capture the corresponding information~\citep{liu2023zero,shi2023mvdream,liu2023syncdreamer}.

\begin{figure}[!t]
\centering
\includegraphics[width=0.99\linewidth]{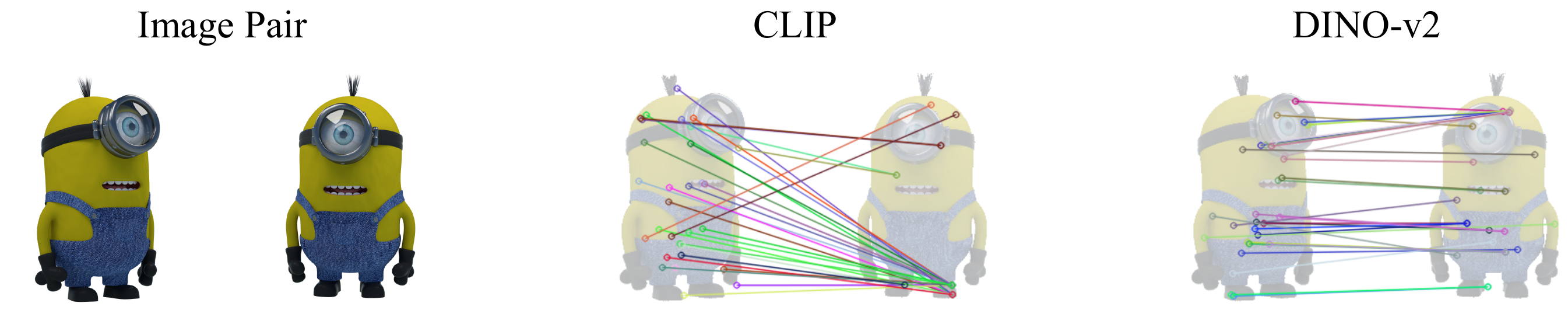}
\caption{\textbf{Visualization of Patch Matching.} We conduct patch-level feature-matching between images under different viewpoints. Patches extracted by DINO-v2 produce better feature-matching results. More details can be found on Appendix~\ref{appendix:pca}.}
\label{fig:matching}
\end{figure}

\begin{figure*}[!t]
\begin{center}
    \centering
    \includegraphics[width=0.99\linewidth]{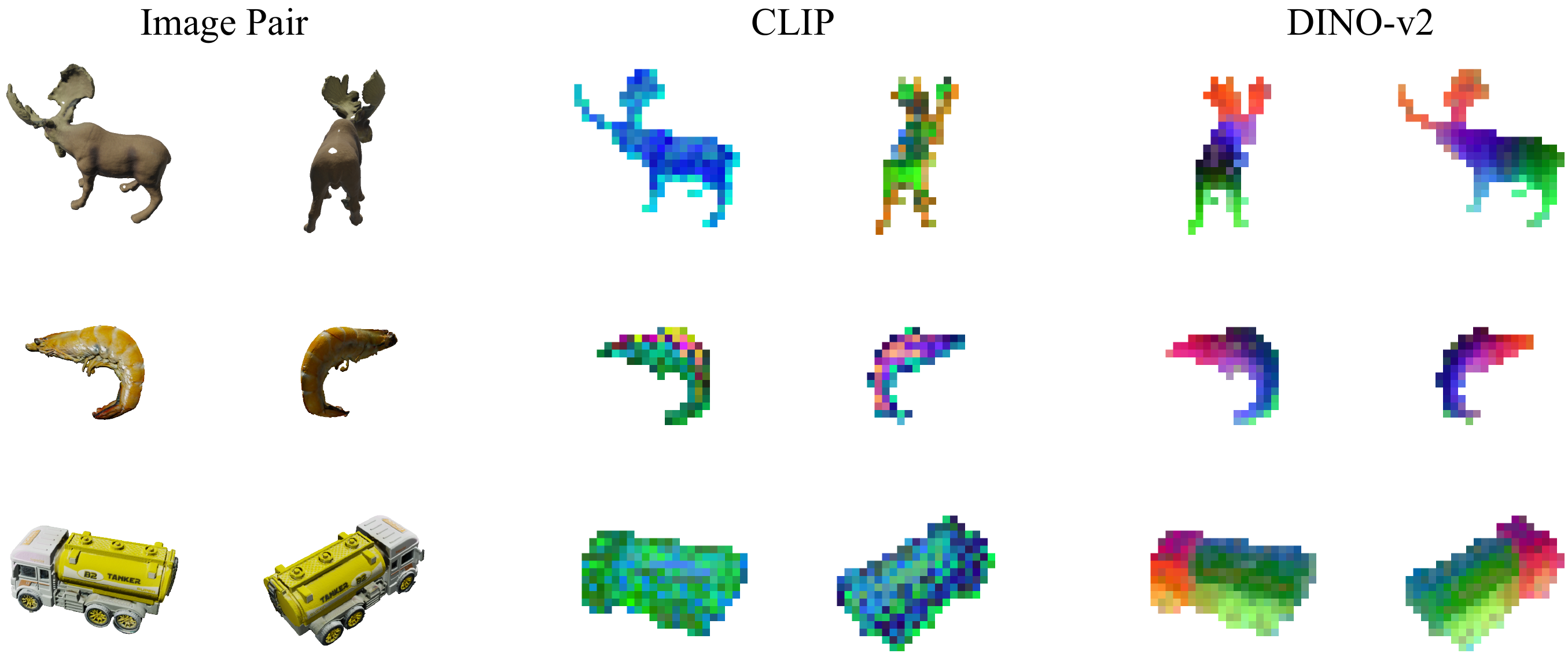} 
\caption{\textbf{Visualization of Principal Component Analysis (PCA).} 
For each object, we use the alpha channel to remove the white background and use PCA to condense the feature dimension to 3 on each view separately. We then visualize the features of patches by normalizing them to 0-255. 
Compared to the CLIP encoder, DINO-v2 exhibits a more coherent semantic understanding of identical regions, due to the fact that the same parts under different viewpoints are matched regardless of poses. More details can be found on Appendix~\ref{appendix:pca}.}
\label{fig:PCA}
\end{center}
\end{figure*}

However, it is noteworthy that the adopted pre-trained Stable Diffusion's foundational training task --- text-to-image generation, exhibits a marginal association with the inference of 3D representations. 
Meanwhile, the CLIP encoder is primarily designed for the purpose of aligning with text embedding, showing poor capability on dense prediction, especially when compared with other image encoders instructed through self-supervised paradigms, E.g., DINO-v2~\citep{oquab2023dinov2}. We show an apple-to-apple comparison between CLIP and DINO-v2 encoder in Figure~\ref{fig:matching} and~\ref{fig:PCA}. Through the Principal Component Analysis (PCA) and correspondence matching visualization, the spatial representations derived via the DINO-v2 encoder show more consistent semantic information than those extracted by the CLIP encoder.

Motivated by this reflection, we advocate for the incorporation of multi-scale representations produced by the DINO-v2 encoder, spanning from the output of its shallow layer to the deeper layer. 
Sequentially, these representations are concatenated and then converted to fit the target channel by a single linear layer. 
After that, we conduct several different spatial interpolation processing, and amalgamate the resultant representations into different stages of UNet's encoder, via an additive operation. 
We also use a cross-attention layer to combine the predicted ``[CLS]'' token with the intermediate feature of the UNet. Details are shown in Figure~\ref{fig:pipeline}.

\subsection{Enhanced Training Paradigm}
Integrating the above methodologies, we accelerate the training process by a large margin. Building on top of this, we continually introduce several engineering enhancements to optimize our framework, culminating in its most efficient version. To start with, we transition from the standard full-precision training scheme to 16-bit mixed-precision scheme, which helps save about $40\%$ training time but leads to numerical error and causes instability. This is mitigated by adding another Layer Normalization~\citep{ba2016layer} before sending the DINO-v2 feature to the diffusion model. Simultaneously, outputs from the DINO-v2 encoder are archived on the disk, given that DINO-v2 does not participate in the backpropagation chain, resulting in a $10\%$ training time reduction further. Lastly, we replace the formal constant learning rate scheduler with a cosine decayed strategy.
As a result of these adjustments, we achieved a total of $50\%$ reduction further in training time, without performance degradation. Detailed training hyperparameters are included in Appendix~\ref{appendix:training}.

\section{Experiments}
\subsection{Dataset}
In our study, we employ the recently released Objaverse~\citep{deitke2023objaverse} dataset for the training processes. The Objaverse dataset contains 800k three-dimensional objects, meticulously crafted by an assemblage of over 100k artists. We adopt 792k samples as the training set and use other 8k samples for validation. 
For the purposes of our methodology, each individual object within the dataset underwent a procedure where 12 viewpoints are samples. To make a fair comparison, we directly adopt the resultant renderings produced by~\cite{liu2023zero}.
During the training phase, a pair of views with $256 \times 256$ resolution is sampled from each object to construct an image duo, designated as ($x$, $\widehat{x}$). We randomly select one as the conditional reference and the other as the ground truth of the predicted novel view. From this duo, the associated relative viewpoint transformation is represented as (R, T).

\begin{figure}[!t]
\vspace{-2em}
    \begin{tabular}{cc}
       \hspace{-1em} \includegraphics[width=0.45\linewidth]{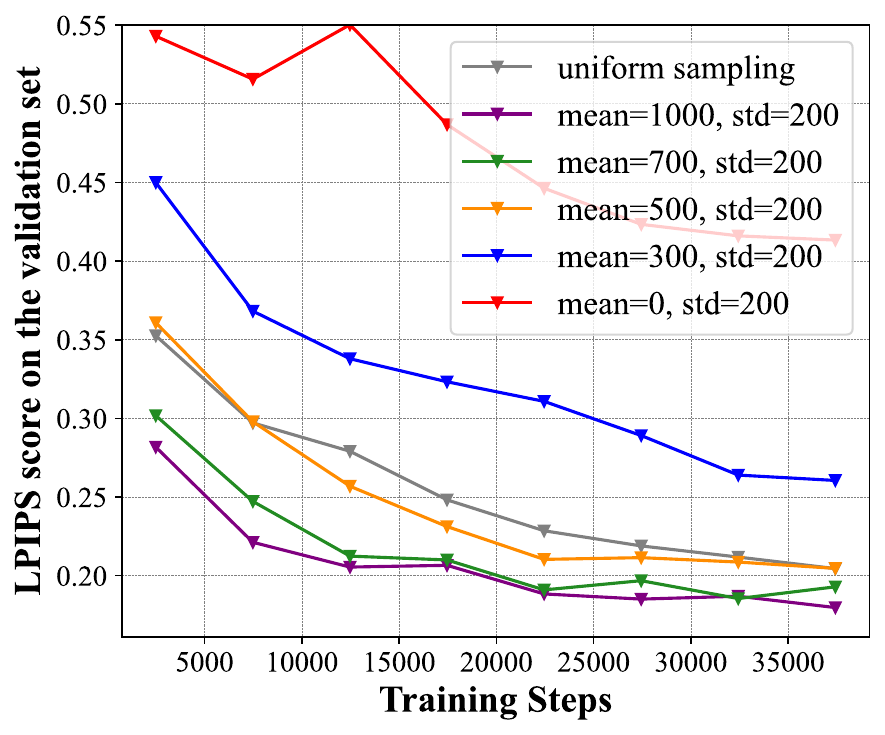}  & \hspace{2em}\includegraphics[width=0.45\linewidth]{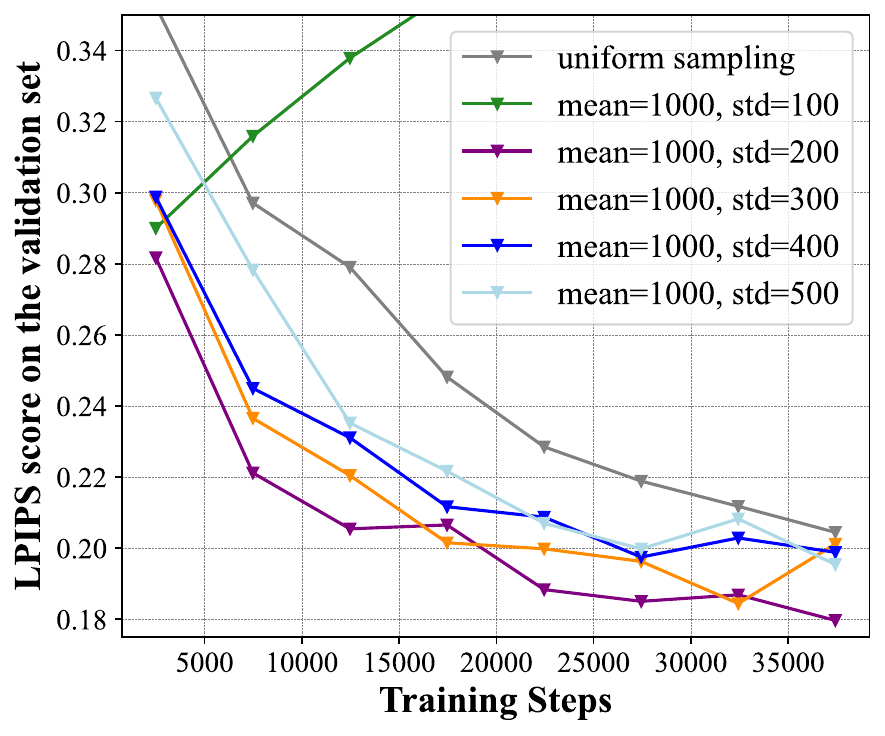} \\
    \end{tabular}
\vspace{-1.em}
\caption{\textbf{Ablation study on different choices of $mean$ and $std$.} The left figure shows different settings by changing the $mean$ factor, and the right figure shows the results by changing the value of $std$. We report the LPIPS scores on the Objaverse validation set for all approaches.}
\label{tab:gaussian_ablation}
\vspace{-2.em}
\end{figure}

\subsection{Quantitative Evaluation}
\subsubsection{Ablation on Different Sampling Strategies}
\label{sec:exp_gaussian}
\vspace{-0.5em}
As discussed in Section~\ref{sec:sampling_strategy}, we are motivated to switch from the standard uniform sampling of diffusion timestep $t$ to a modified sampling strategy that allows us to conserve efforts on superfluous segments of the training. Our initial assumption is built upon the observation from the diffusion model's inference process, where the noisy stage should receive augmented opportunities for updates. To verify it, we conduct experiments by controlling the value of mean and standard deviation. We first fix the standard deviation to be $200$ and switch the value of mean from a group of \{$1000$, $700$, $500$, $300$, $0$\}, as shown in Figure~\ref{fig:gaussian} left. 
After that, we keep the value of mean to be $1000$ and test several options for standard deviation, from a group of \{$100 \sim 500$\}, as depicted in Figure~\ref{fig:gaussian} right. The results demonstrate our assumption --- the emergence of novel view synthesis ability is more correlated with the late stage of diffusion steps (which takes a more noisy input). Meanwhile, the results also suggest that other stages should be still updated at a reduced frequency, rather than entirely omitted. For instance, in the setting of $mean = 1000$ and $std=100$, the early stage of diffusion steps have almost zero probability of being updated, and the resultant framework shows worse results on the validation set.

\subsubsection{Ablation Study on Different Components}
\begin{wraptable}{r}{7.7cm}
\vspace{-2.em}
\caption{Ablation study on different components. We train each method 20,000 steps and then report the LPIPS and MSE score on the Objaverse validation set.}
\label{tab:ablation}
\vspace{0.5em}
\begin{tabular}{C{1cm}C{0.7cm}C{1.9cm}C{1cm}C{1cm}}
    \toprule
         Gaussian&DINO&Amalgamation & LPIPS$\downarrow$ & MSE$\downarrow$ \\
    \hline
         \usym{2717} & \usym{2717} & \usym{2717} & 0.236 & 0.187  \\
         \Checkmark  & \usym{2717} & \usym{2717} & 0.205 & 0.142 \\
         \Checkmark  & \Checkmark  & \usym{2717} & 0.187 & 0.136 \\
         \Checkmark  & \usym{2717} & \Checkmark  & 0.191 & 0.144 \\
         \Checkmark  & \Checkmark  & \Checkmark  & \textbf{0.171} & \textbf{0.128} \\
    \bottomrule
    \end{tabular}
    \vspace{-1.em}
\end{wraptable} 
We conduct an ablation study to demonstrate the effectiveness of our proposed components, as shown in Table~\ref{tab:ablation}. The official checkpoint of the baseline method, Zero 1-to-3~\cite{liu2023zero}, requires \textbf{105,000} training iterations. Contrarily, in our experimental configuration, we constrict each counterpart to a resource-limited training scenario, allowing only \textbf{20,000} updates. Given these constraints, the baseline approach exhibits subpar performance on the validation set, only reaching 0.279 LPIPS score and 0.212 Mean Squared Error (MSE). By replacing standard uniform sampling with Gaussian sampling \{$mean = 1000$, $std=200$\}, we observe improvements in both the LPIPS and MSE metrics. Next, we attempt to replace the CLIP encoder with the DINO-v2 encoder, or amalgamate the spatial feature from these encoders with the UNet denoiser. Either of them brings performance gain. Finally, we add all these modifications to the framework, reaching the best performance by obtaining a 0.171 LPIPS score and 0.128 MSE score on the validation set.

\begin{figure*}[!t]
\vspace{-2.em}
\begin{center}
    \centering
    \includegraphics[width=0.9\linewidth]{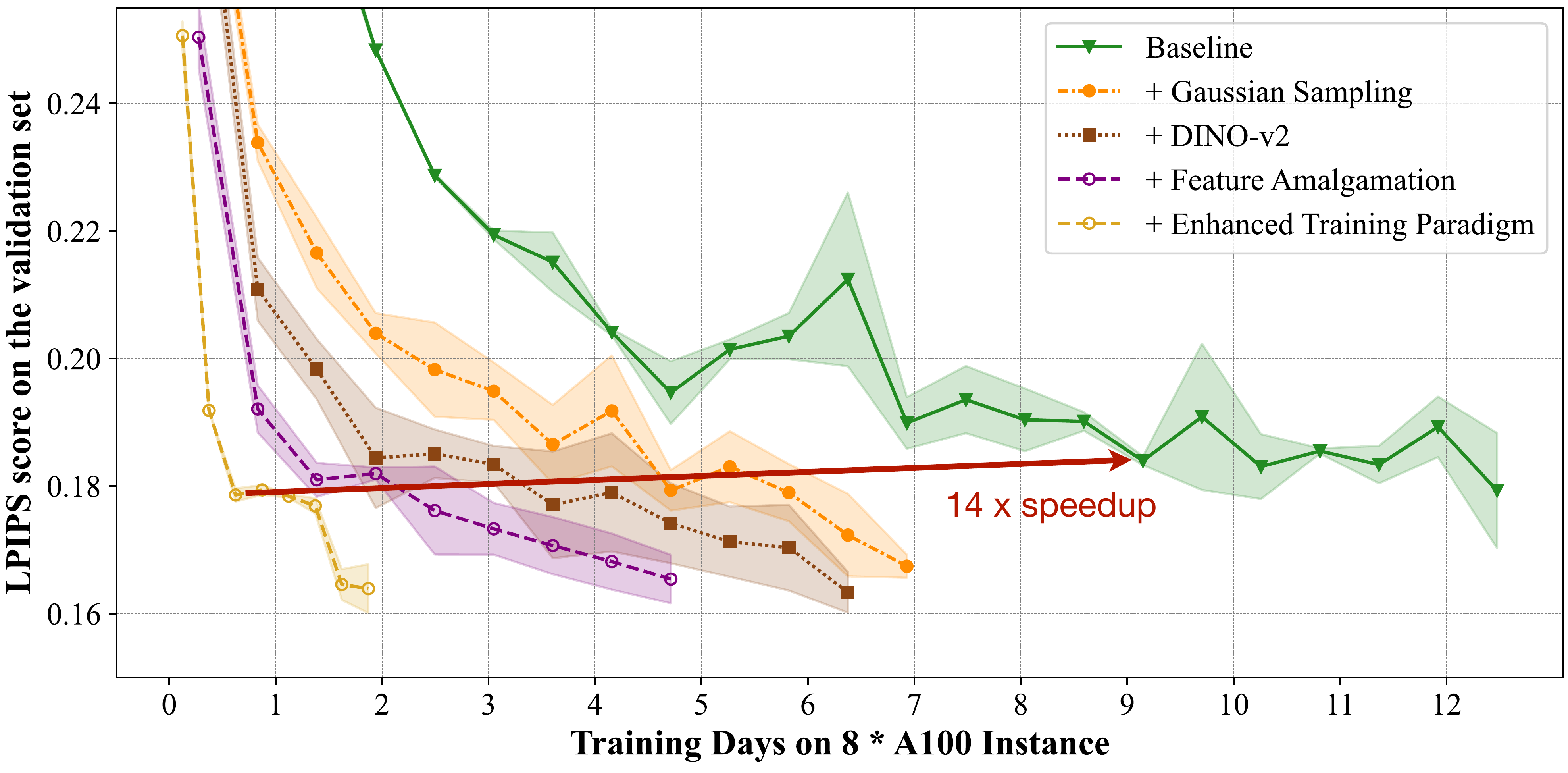} 
\vspace{-1.em}
\caption{\textbf{Ablation study on different settings.} To verify the effectiveness of different components in our framework, we conduct three experiments for each setting with different random seeds and report the LPIPS score~\citep{zhang2018unreasonable} on the Objaverse validation set. }
\label{fig:benchmark}
\end{center}
\end{figure*}

Meanwhile, we present the curve of validation loss w.r.t. the training time, shown in Figure~\ref{fig:benchmark}. Built upon the aforementioned best setting (\textcolor{violet}{purple} line), we further apply the proposed enhanced training paradigm (\textcolor{brown}{brown} line). Compared to the Zero 1-to-3 baseline, the final resultant framework reaches up to $14\times$ speedup.

\subsection{Visual Comparison on In-the-wild Images}
\label{sec:in-the-wild}
\begin{figure*}[!t]
\vspace{-1em}
\begin{center}
    \centering
    \includegraphics[width=0.99\linewidth]{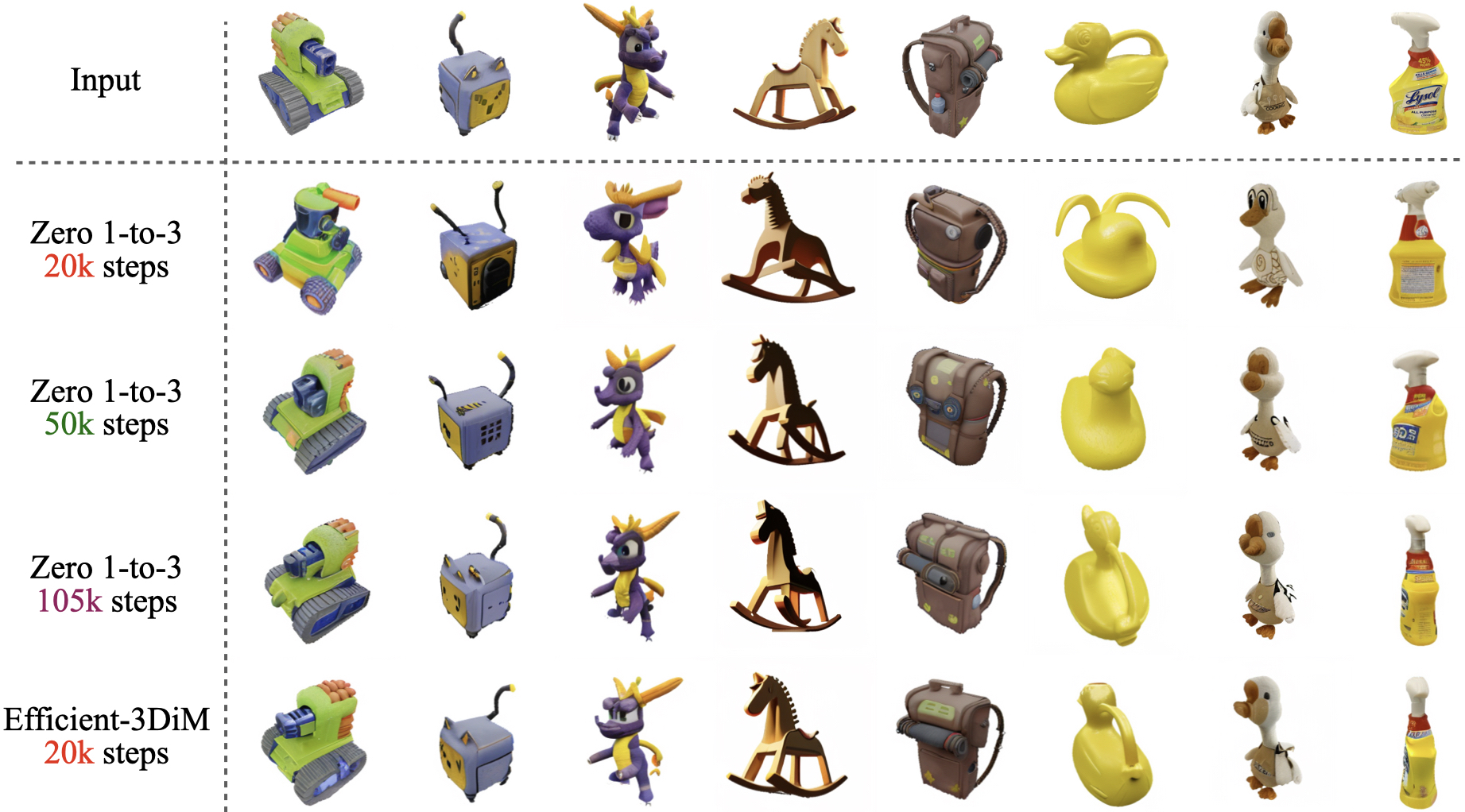} 
\vspace{-1.em}
\caption{Visluazation comparisons on in-the-wild images. For every object, we turn the camera 90 degrees counterclockwise around the azimuth axis.}
\label{fig:visual_comparison}
\end{center}
\vspace{-2em}
\end{figure*}

\vspace{-0.5em}
We examine the performance of the obtained model on in-the-wild images. As shown in Figure~\ref{fig:visual_comparison}, if we reduce the training iterations of Zero 1-to-3 from the original 105k steps to only 200k or 500k steps, it fails to produce multi-view consistent visual outputs. However, the proposed method can generate photorealistic and geometrically reasonable novel views under the same training steps.
\vspace{-0.5em}

\section{Conclusion and Future Works}
\vspace{-0.5em}
In this paper, we introduce Efficient-3DiM, an efficient framework for single-image novel view synthesis through diffusion models. At its core, Efficient-3DiM integrates three pivotal contributions: a modified sampling strategy departing from traditional uniform sampling, an integration of a self-supervised Vision Transformer replacing the conventional CLIP encoder, and an enhanced training paradigm. 
Comprehensive evaluations underscore the effectiveness of proposed components, successfully slashing training time from 10 days to a single day, all the while preserving computational resource integrity. As with other approaches, there remain avenues for exploration and improvement. 
One such avenue could involve further refinement of the multi-view consistency, ensuring not only efficiency but also better fidelity. Going forward, we will explore further ideas on this direction.

\section{Acknowledgement}
We thank our colleagues at Apple, especially Wenze Hu, Haotian Zhang, and Zhe Gan, for their valuable support in enhancing the quality of this work.
\bibliography{iclr2024_conference}
\bibliographystyle{iclr2024_conference}
\appendix
\section{Appendix}
\begin{figure*}[!b]
\begin{center}
    \centering
    \includegraphics[width=0.99\linewidth]{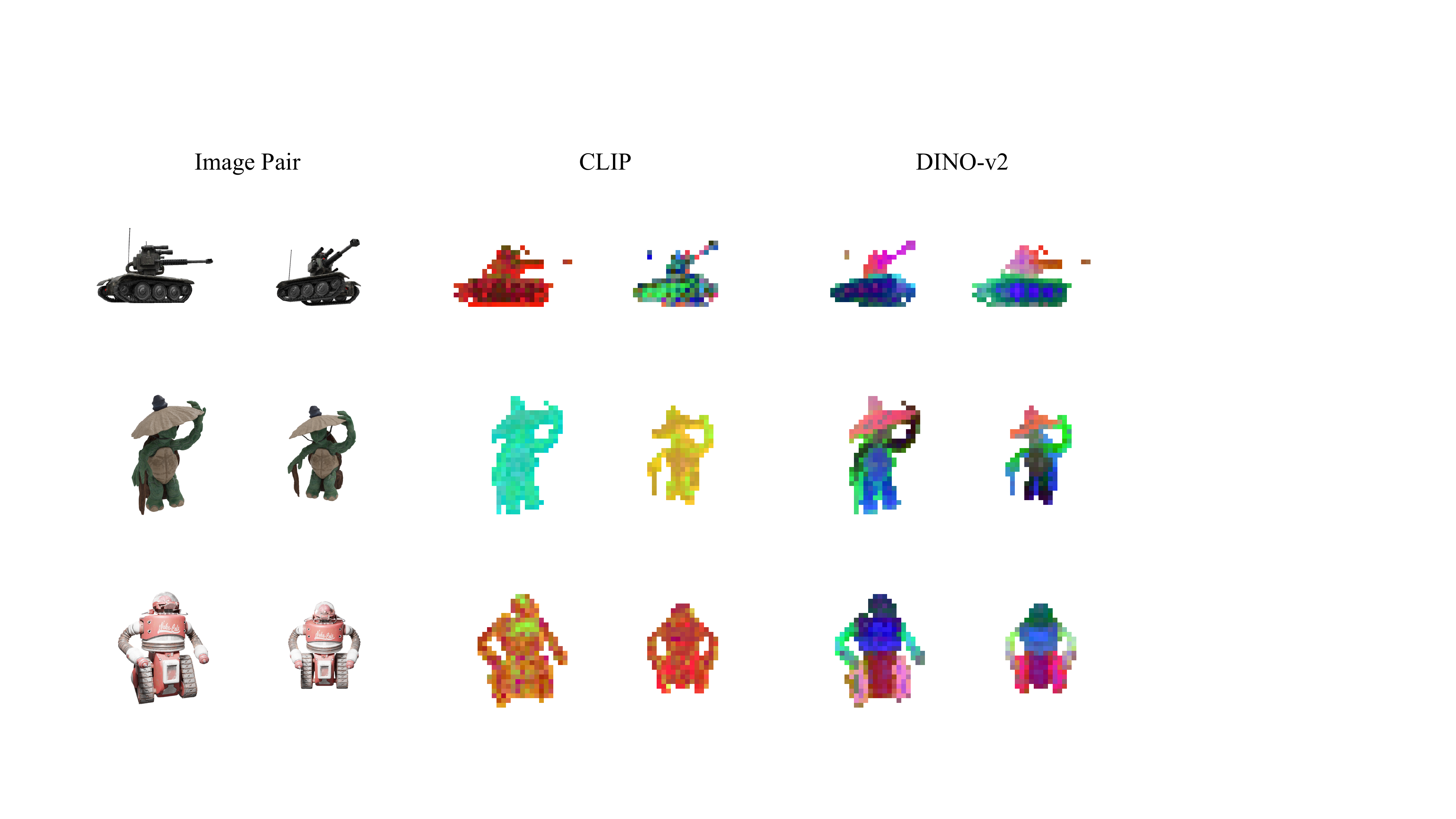} 
\caption{Visualization of Principal Component Analysis (PCA).}
\label{fig:supp_pca}
\end{center}
\vspace{-1em}
\end{figure*}
\subsection{Detailed Training Scheme}
\label{appendix:training}
All of our experiments are conducted on 8 Nvidia-A100 GPUs using the PyTorch-Lightning-1.4.2 platform. We apply a batch size of 48 per GPU and adopt gradients accumulation by 4 times. Thus the real batch size is 192 * 8 in total. We adopt an Adam optimizer with $\beta_1 = 0.9$, $\beta_2 = 0.999$, and $0.01$ weight decay.  We adopt a half-period cosine schedule of learning rate decaying with the base learning rate to be $1e-4$, the final learning to be $1e-5$, and the maximum training iterations set to be 30,000. A linear warm-up strategy is applied to the first 200 steps. We also use exponential moving average weights for the UNet denoiser. We select ``ViT-L/14'' for both the CLIP encoder and the DINO-v2 encoder for a fair comparison. The in-the-wild images tested on Section~\ref{sec:in-the-wild} are taken from~\citep{liu2023one}.

\subsection{Principal Component Analysis and Feature-matching Visualization}
For Principal Component Analysis (PCA) visualization, we first use the selected encoder to capture the spatial feature maps from two viewpoints of each object. Subsequently, we employ the alpha channel to eliminate the white background. This results in $16 \times 16$ patches with $c$ channels. After that, we separately compute the Principal Component Analysis (PCA) of these two features and select the first three components. Each component is assigned a color channel, where we are able to further visualize it by using RGB color space. For feature-matching visualization, we directly adopt the condensed feature maps from previously conducted PCA procedures. We compute the Euclidean distance between patches extracted from two viewpoints and then map each of them. More visual results of PCA can be found in Figure~\ref{fig:supp_pca}



\label{appendix:pca}

\end{document}